\documentclass[sigconf]{acmart}
\pdfoutput=1
\usepackage{amsmath}
\usepackage{multirow}
\usepackage{array}
\usepackage{balance}
\AtBeginDocument{%
  \providecommand\BibTeX{{%
    \normalfont B\kern-0.5em{\scshape i\kern-0.25em b}\kern-0.8em\TeX}}}

\setcopyright{acmcopyright}

\copyrightyear{2020}
\acmYear{2020}
\setcopyright{acmcopyright}\acmConference[MM '20]{Proceedings of the 28th ACM International Conference on Multimedia}{October 12--16, 2020}{Seattle, WA, USA}
\acmBooktitle{Proceedings of the 28th ACM International Conference on Multimedia (MM '20), October 12--16, 2020, Seattle, WA, USA}
\acmPrice{15.00}
\acmDOI{10.1145/3394171.3413904}
\acmISBN{978-1-4503-7988-5/20/10}

\begin{document}
\fancyhead{}

\title{Domain Adaptive Person Re-Identification via Coupling Optimization}

\author{Xiaobin Liu, Shiliang Zhang}
\affiliation{ School of Electronics Engineering and Computer Science, Peking University \\ \{xbliu.vmc, slzhang.jdl\}@pku.edu.cn}

\begin{abstract}
  Domain adaptive person Re-Identification (ReID) is challenging owing to the domain gap and shortage of annotations on target scenarios. To handle those two challenges, this paper proposes a coupling optimization method including the Domain-Invariant Mapping (DIM) method and the Global-Local distance Optimization (GLO), respectively. Different from previous methods that transfer knowledge in two stages, the DIM achieves a more efficient one-stage knowledge transfer by mapping images in labeled and unlabeled datasets to a shared feature space. GLO is designed to train the ReID model with unsupervised setting on the target domain. Instead of relying on existing optimization strategies designed for supervised training, GLO involves more images in distance optimization, and achieves better robustness to noisy label prediction. GLO also integrates distance optimizations in both the global dataset and local training batch, thus exhibits better training efficiency. Extensive experiments on three large-scale datasets, \textit{i.e.}, \textit{Market-1501}, \textit{DukeMTMC-reID}, and \textit{MSMT17}, show that our coupling optimization outperforms state-of-the-art methods by a large margin. Our method also works well in unsupervised training, and even outperforms several recent domain adaptive methods.

\end{abstract}




\maketitle

\section{Introduction}
Person Re-Identification (ReID)~\cite{wei2018vp, li2019global, Chen_2019_ICCV_attention, zha2020adversarial, huang2020real, wang2020unsupervised, jianing_eccv_2020, zhong2020robust, li2020multi, liu2019self} targets to match a query person image against a gallery set. Recent works have achieved promising performance in supervised scenario~\cite{Chen_2019_ICCV_attention, zhang2017alignedreid, Guo_2019_ICCV, Liu_2019_ICCV, Zhang_2019_CVPR}. However, supervised ReID models suffer from the expensive data annotation and substantial performance drop when applied on different target domains. To address those issues, recent works focus on domain adaptive person ReID by transferring knowledge learned from the labeled domain to the target domain~\cite{yang2019patch, zhong2019invariance, yu2019unsupervised, Liu_2019_CVPR}. Transferred models exhibit better generalization ability and can be applied on target domains without labeled data. More details of related works are summarized in Sec.~\ref{sec:related_work}.

\begin{figure}[t]
\begin{center}
\includegraphics[width=0.99\linewidth]{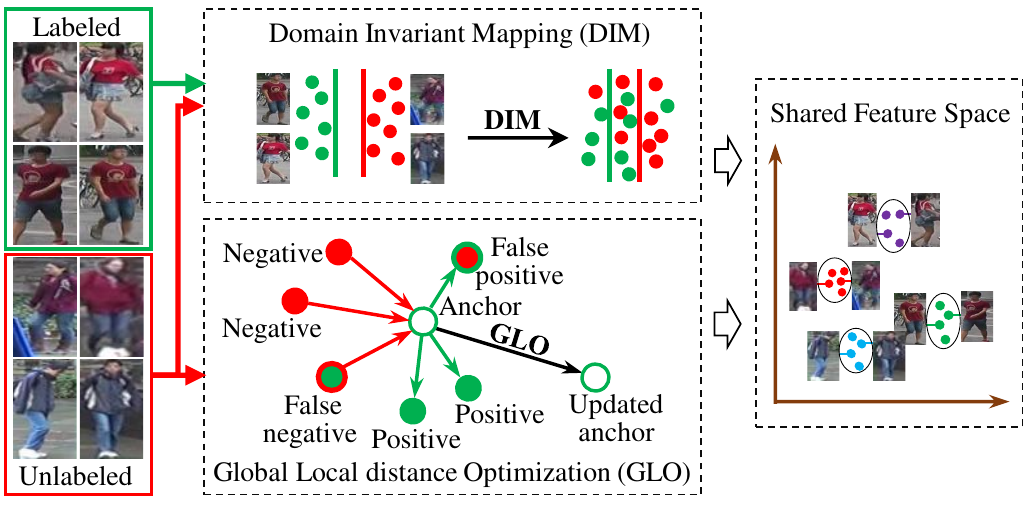}
\end{center}
\vspace{-5mm}
\caption{Illustration of proposed coupling optimization method consisting of Domain-Invariant Mapping (DIM) and Global-Local distance Optimization (GLO). In DIM, green and red dots denote labeled and unlabeled samples, respectively. In GLO, red, green, and black arrows denote directions of push, pull and update for anchor, respectively. In the shared feature space, dots in different colors denote samples of different identities.}
\label{fig:intro}
\vspace{-4mm}
\end{figure}

Despite the significant success, domain adaptive person ReID is still a challenging task and there remain several open issues unexplored. Firstly, previous works commonly conduct the knowledge transfer in two stages, \emph{i.e.}, first transfer labeled images to the target domain with Generative Adversarial Networks (GANs), then train the ReID model~\cite{Chen_2019_ICCV, Liu_2019_CVPR, huang2019sbsgan, Li_2019_ICCV} using transferred images. However, GANs could be hard to tune.The image generation is also challenging and sensitive to various factors like backgrounds, lighting, \emph{etc.}. Secondly, person ReID models for the unlabeled target domain can be optimized by predicted labels. This procedure requires an optimization strategy robust to noisy labels. Existing unsupervised person ReID works commonly adopt the optimization strategies in supervised ReID training, \emph{e.g.}, triplet loss~\cite{yu2019unsupervised, zhong2018generalizing, ssg}, which are not specifically optimized for unsupervised training.

This paper is motivated to study an efficient optimization for domain adaptive person ReID. As shown in Fig.~\ref{fig:intro}, we first propose an efficient one-stage knowledge transfer model. It trains the target ReID model using labeled source dataset and unlabeled target dataset, without requiring image transfer and GANs training. An efficient unsupervised optimization is further proposed to conduct the optimization on the target domain. Those two optimizations jointly boost the performance of domain adaptive person ReID.

As shown in Fig.~\ref{fig:intro}, the DIM method is proposed to effectively transfer discriminative cues from the labeled dataset to the target dataset. DIM bridges the domain gap between labeled and unlabeled datasets in feature space in an adversarial learning manner. Specifically, a domain discriminator is trained based on the learned person ReID feature to discriminate where samples are from. The feature extractor is optimized to confuse the domain discriminator. Those two models are interactively optimized, leading to a discriminative feature space shared by both the labeled and unlabeled datasets. The DIM directly bridges the domain gap in feature space, hence is more efficient than the two-stage transfer models.

Individual image triplets or pairs contain a small number of images, making triplet loss sensitive to errors in label prediction. As illustrated in Fig.~\ref{fig:intro}, to enhance the robustness to noisy labels, one possible way is to involve more images during distance optimization. This intuition leads to the GLO strategy. GLO consists of Global Optimization (GO) and Local Optimization (LO), both compute the distance among a larger number of samples. GO optimizes the distance of each anchor image and its positive and negative images in the training set. GO is achieved with the memory bank~\cite{wu2018unsupervised}, which caches the features of the entire training set. LO is introduced for distance optimization in each training batch. Note that, each training batch samples a small portion of training set. This sampling strategy guarantees images inside each training batch likely to be from different persons. The LO is hence computed by pushing all images away from each other. Compared with GO, the LO is more efficient to compute and accelerates the training convergence.

Our person ReID model is jointly optimized with the above two training algorithms. We hence call our method as coupling optimization. We test our model on three large-scale datasets, \textit{i.e.}, \textit{Market-1501}~\cite{market}, \textit{DukeMTMC-reID}~\cite{duke}, and \textit{MSMT17}~\cite{msmt}. Comparison with recent works shows that our methods outperform state-of-the-art works by a large margin. For instance, using \textit{DukeMTMC-reID} as the source domain, our method achieves Rank-1 accuracy of 88.3\% on \textit{Market-1501}, outperforming recent PAST~\cite{Zhang_2019_ICCV} and SSG~\cite{ssg} by 9.9\% and 8.3\%, respectively. We also test our method in unsupervised scenario, \emph{i.e.}, training the ReID model only with GLO. Our method achieves Rank-1 accuracy of 77.4\% on \textit{Market-1501}, significantly outperforming recent BUC~\cite{lin2019bottom} and DBC~\cite{ding2019towards} by 11.2\% and 8.2\%, respectively. It is worth noting that, our unsupervised training also outperforms several domain adaptive methods that use extra source domain for training, such as PAUL~\cite{yang2019patch} and DA\_2S~\cite{huang2019sbsgan}.

In summary, this work iteratively runs DIM and GLO to optimize the domain adaptive person ReID model. DIM transfers discriminative cues from the source domain to the target domain. GLO achieves unsupervised optimization based on predicted labels on the target domain. Compared with existing works, DIM and GLO exhibit advantages in efficiency and robustness to label noises, respectively. Those properties guarantee the superior performance of our method in comparisons with state-of-the-art works.

\section{Related Work}
\label{sec:related_work}
This work is closely related with unsupervised domain adaptation, supervised and domain adaptive person ReID. This section briefly reviews those three categories of works.

\subsection{Unsupervised Domain Adaptation}
Unsupervised domain adaptation (UDA) is similar to transfer learning for person ReID. However, UDA holds a strong assumption that the unlabeled domain has the same categories with the labeled domain~\cite{wang2019transferable, yang2019cross}. While in domain adaptive person ReID, labeled and unlabeled domains usually have totally different identities. The same issue that both tasks suffer from is the domain gap. Several works on UDA have been proposed to tackle this issue~\cite{ganin2015unsupervised, kumagai2019unsupervised, long2015learning,  tzeng2017adversarial}. Ganin \textit{et al.}~\cite{ganin2015unsupervised} propose a gradient reversal layer to learn domain-invariant features, which suffers gradient vanish issue in training. Kumagai \textit{et al.}~\cite{kumagai2019unsupervised} and Long \textit{et al.}~\cite{long2015learning} propose to transform the feature distribution of labeled domain to approach the feature distribution of unlabeled domain. However, those methods may suffer from the noisy from mean feature computation. Moreover, encouraging the distributions of labeled and unlabeled domains to be identical is not reasonable for person ReID task because different domains have different identities. For instance, colors of clothe in unlabeled domain could be different from labeled domain.
Tzeng \textit{et al.}~\cite{tzeng2017adversarial} present an adversarial loss to encourage the feature domain of unlabeled dataset to approach the one of labeled dataset. However, this restriction is possible to disturb the distance optimization on unlabeled dataset. Different from~\cite{tzeng2017adversarial}, proposed DIM restricts features from both labeled and unlabeled datasets to approach to each other, which alleviates the disturbance on distance optimization and is also more effective at narrowing domain gap.

\subsection{Supervised Person Re-Identification}
Supervised person ReID task has been widely studied and many methods have been proposed from many respects, such as parts feature extraction~\cite{dlpar, spindle, longhui, jianing, li2019pose, liu2018ram}, distance metric learning~\cite{zhou2017point, quadruplet, ldns, defense, liu2019group}, attention learning~\cite{Chen_2019_ICCV_attention, teng2018scan}, \textit{etc.} However, manual person ID annotation is expensive and rarely available in real-world application. Although these supervised methods have achieved superior performance on existing labeled datasets, person ReID in unsupervised scenario still remains challenging. Different from these works, this paper focuses on the domain adaptive person ReID task, which is more challenging and also more valuable for real-world applications.

\subsection{Domain Adaptive Person Re-Identification}
Recently, many works have been proposed for domain adaptive person ReID task from different aspects. Some works targeting to narrow domain gaps in image space by GANs~\cite{msmt, zhong2018generalizing, Chen_2019_ICCV, huang2019sbsgan, Liu_2019_CVPR, Li_2019_ICCV}. For example, Liu \textit{et al.}~\cite{Liu_2019_CVPR} propose an adaptive transfer network to effectively transfer images from label domain to unlabeled domain. However, transfer in image space cannot guarantee the elimination of domain gaps in feature space, and the quality of generated images largely affects the performance.
Several works try to reduce domain gaps in feature space~\cite{ huang2019domain, ganin2016domain}. Ganin \textit{et al.}~\cite{ganin2016domain} reverse the gradient of domain discriminator to train the feature extractor. However, the scale of gradient is small when discriminator is well-trained, leading to an ineffective learning. Huang \textit{et al.}~\cite{huang2019domain} encourage the discriminator to have the same output for features from both labeled and unlabeled domains. However, this will end up with the discriminator outputting a same value for all inputs, instead of narrowing down the gap.
Experiments in Sec.~\ref{sec:model_analysis} demonstrate the advantages of proposed DIM against previous methods.

Some researchers focus on unsupervised optimization on unlabeled dataset. Huang \textit{et al.}~\cite{huang2020real} propose a degradation invariance learning approach with depredated surveillance data. It simultaneously extracts robust features and removes real-world degradations without extra supervision, leading to good robustness and effectiveness. Some works locally predict labels for model training~\cite{zhong2019invariance, yu2019unsupervised}. For example, Yu \textit{et al.}~\cite{yu2019unsupervised} select positive samples in training batches. However, local label prediction is not precise and will mislead the distance optimization.
Based on assigned labels, some works adopt triplet loss for model training~\cite{adaptive-reid, ssg, Zhang_2019_ICCV, yu2019unsupervised}. However, these optimization methods fail to consider the noise in predicted labels on unlabeled dataset. Compared with these methods, proposed GLO method is more robust to noisy label and optimizes distance relationship more effectively.

Some works adopt additional cues to boost performance~\cite{Qi_2019_ICCV, li2019unsupervised-pami, li2018unsupervised,ssg, Zhang_2019_ICCV}. For example, Qi \textit{et al.}~\cite{Qi_2019_ICCV} use temporal information and
Li \textit{et al.}~\cite{li2019unsupervised-pami, li2018unsupervised} adopt tracklet information to predict more precise labels on unlabeled datasets. Fu \textit{et al.}~\cite{ssg} and Zhang \textit{et al.}~\cite{Zhang_2019_ICCV} use local features to improve the performance. 
Experimental results show that our model outperforms these methods by a clear margin even without additional cues.


\section{Problem Formulation}
Given a labeled dataset $\mathcal{S}$ and a target unlabeled dataset $\mathcal{T}$, domain adaptive person ReID aims to train the ReID model with both $\mathcal{S}$ and $\mathcal{T}$, and guarantee its discriminative power on $\mathcal{T}$. $\mathcal{S}$ and $\mathcal{T}$ can be denoted as $\mathcal{S}=\{ x_i, y_i | i = 1...N_\mathcal{S}, y_i \in \{1,...,{M}\} \}$ and $\mathcal{T}=\{ t_i | i = 1...N_\mathcal{T}\}$, respectively. $x_i$, $y_i$, $N_\mathcal{S}$, and ${M}$ are the $i$-th image, its label, the number of images and identities in $\mathcal{S}$, respectively. $t_i$ and $N_\mathcal{T}$ denote the the $i$-th image and the number of images in $\mathcal{T}$. We denote the feature extraction as $f_i = {\Phi}(x_i, \theta)$ and $v_i = \Phi(t_i, \theta)$ on $\mathcal{S}$ and $\mathcal{T}$, respectively. $\Phi$ and $\theta$ denote the CNN feature extractor and its parameters, respectively.

The target of model training is to make extracted features discriminative for person ReID task on $\mathcal{T}$. We achieve this goal from two aspects: 1) transfer the discriminative cues learned from $\mathcal{S}$ to $\mathcal{T}$, and 2) optimize the ReID model on $\mathcal{T}$ through unsupervised training. As person ID labels are available in $\mathcal{S}$, cross entropy loss $\mathcal{L}_{CE}$ can be computed on $\mathcal{S}$ to learn discriminative features. To bridge the domain gap between $\mathcal{T}$ and $\mathcal{S}$, DIM maps $\mathcal{T}$ and $\mathcal{S}$ into a shared feature space. To achieve unsupervised training, we first predict person ID labels on $\mathcal{T}$, then optimize the ReID model with GLO. Therefore, our ReID model is optimized by three training losses, \emph{i.e.}, the cross entropy loss $\mathcal{S}$, the DIM loss on $\mathcal{S}$ and $\mathcal{T}$, as well as the GLO loss on $\mathcal{S}$. The overall loss function can be formulated as:
\begin{eqnarray}
\mathcal{L} = \mathcal{L}_{CE} + \lambda_{DIM} \mathcal{L}_{DIM} + \mathcal{L}_{GLO},
\end{eqnarray}
where $\mathcal{L}_{CE}, \mathcal{L}_{DIM}, \mathcal{L}_{GLO}$ denote the three losses discussed above. $\lambda_{DIM}$ is loss weights.

$\mathcal{L}_{DIM}$ is expected to map $\mathcal{T}$ and $\mathcal{S}$ into a shared feature space, where features from different datasets show similar distribution. This shared feature space hence could be optimized by both $\mathcal{T}$ and $\mathcal{S}$, \emph{i.e.}, it can be optimized by $\mathcal{L}_{CE}$ on $\mathcal{S}$ and optimized by $\lambda_{GLO}$ on $\mathcal{T}$ to gain better discriminative power. To learn this shared feature space, DIM first trains a discriminator network DNet, which is able to discriminate the source domain of each sample based on its feature. The CNN feature extractor $\Phi$ is updated to confuse DNet. This adversarial training strategy iteratively updates DNet and $\Phi$, finally achieves a discriminative feature space shared by both $\mathcal{T}$ and $\mathcal{S}$. Compared with existing two-stage image transfer models, DIM does not involve GANs training and image transfer, thus efficiently bridges the domain gap between $\mathcal{T}$ and $\mathcal{S}$. More details of DIM computation could be found in Sec.~\ref{sec:dim}.

$\mathcal{L}_{GLO}$ is computed on $\mathcal{T}$ based on predicted labels. Label prediction can be conducted on the entire $\mathcal{T}$ and inside each training batch, respectively. Those two predictions lead to the Global Optimization (GO) and Local Optimization (LO), respectively. GO first generates image clusters on $\mathcal{T}$, then introduces a distance threshold to select positive pairs inside each cluster. To further eliminate the negative effects of noisy label prediction, GO involves more samples for loss computation. More specifically, for each anchor image GO generates a group of positive samples and a group of negative samples, then optimizes the group distance, instead of the sample distance. As shown in Fig.~\ref{fig:optimization}, this strategy leads to better robustness to label noises than triplet loss. LO is conducted based on label prediction inside each training batch. Each training batch randomly samples a small number of images from the large-scale $\mathcal{T}$. This strategy selects images with different ID labels with high probability. The LO is hence computed to push samples in the same training batch far away from each other.

We hence denote $\mathcal{L}_{GLO}$ as the combination of GO and LO, \emph{i.e.}, 
\begin{eqnarray}
\mathcal{L}_{GLO} = \lambda_{GO} \mathcal{L}_{GO}+ \lambda_{LO} \mathcal{L}_{LO},
\end{eqnarray}
where $\lambda_{GO}$ and $\lambda_{LO}$ are loss weights. $\mathcal{L}_{GO}$ and $\mathcal{L}_{LO}$ are both computed to optimize the ReID model on $\mathcal{T}$. Compared with $\mathcal{L}_{GO}$, $\mathcal{L}_{LO}$ is more efficient in label prediction and loss computation. As shown in experiments, $\mathcal{L}_{LO}$ substantially accelerates the training convergence. Details of computation to $\mathcal{L}_{GLO}$ will be given in Sec.~\ref{sec:glo}.

\section{Domain-Invariant Mapping}
\label{sec:dim}

Owing to the domain gap between $\mathcal{S}$ and $\mathcal{T}$, even model has been well trained on $\mathcal{S}$, the performance often drops a lot on $\mathcal{T}$. To address this issue, we introduce the DIM method to encourage the model to map images from both $\mathcal{S}$ and $\mathcal{T}$ into a shared domain in feature space. DIM is designed in an adversarial learning manner. Specifically, we use a discriminator network DNet to recognize which domain images come from. DNet takes as input features from both source and target datasets, and output domain recognition scores. We train DNet by Euclidean distance loss with the output target to be 1 for $f_i$ from $\mathcal{S}$ and 0 for $v_i$ from $\mathcal{T}$. This objective can be formulated as follows:
\begin{eqnarray}
\label{eqn:DNet}
\ell_{DNet} = \frac{1}{N_{\mathcal{S}}} \sum_{i=1}^{N_\mathcal{S}} ({\rm DNet}(f_i) - 1)^2 + \frac{1}{N_{\mathcal{T}}} \sum_{i=1}^{N_\mathcal{T}} {\rm DNet}(v_i)^2.
\end{eqnarray}
After training by Eqn.~\eqref{eqn:DNet}, DNet is able to recognize which domain a feature belongs. Then, we use DNet to supervise $\Phi$ with the objective that $\Phi$ can confuse DNet by extracting features whose domain recognition scores are 0.5. 
The objective function to train $\Phi$ can be formulated as:
\begin{eqnarray}
\label{eqn:dim}
\mathcal{L}_{DIM}(\theta) \! = \! \frac{1}{N_\mathcal{S}} \!\! \sum_{i=1}^{N_\mathcal{S}}  ({\rm DNet}(f_i) \!-\! 0.5)^2  \!+\! \frac{1}{N_\mathcal{T}}\!\! \sum_{i=1}^{N_\mathcal{T}} \! ({\rm DNet}(v_i) \!-\! 0.5)^2. \!\!\!
\end{eqnarray}

\begin{figure}[t]
\begin{center}
\includegraphics[width=0.99\linewidth]{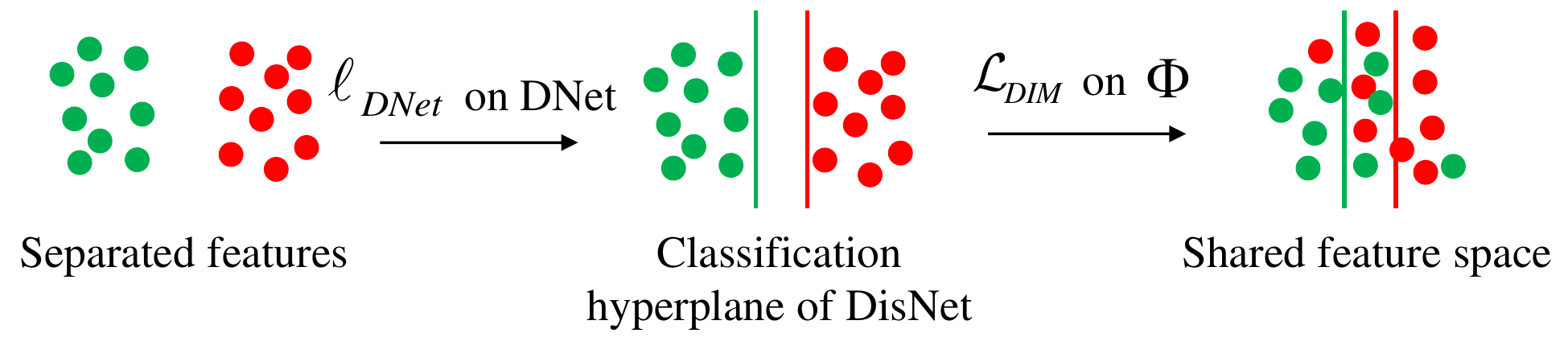}
\end{center}
\vspace{-5mm}
\caption{Illustration of proposed DIM. Dots in different colors are of different domains.}
\label{fig:dim}
\vspace{-4mm}
\end{figure}

The training of DNet and $\Phi$ is conducted interactively as illustrated in Fig.~\ref{fig:dim}. In each iteration, we first train DNet with Eqn.~\eqref{eqn:DNet}. Then, we train $\Phi$ with Eqn.~\eqref{eqn:dim} to confuse DNet. By interactively training DNet and $\Phi$, $\Phi$ is finally able to map images from different domains into a shared feature space.

Compared with the method in~\cite{tzeng2017adversarial} that only restricts features from $\mathcal{T}$, $\mathcal{L}_{DIM}$ also pulls features in $\mathcal{S}$ to approach features in $\mathcal{T}$. As ground truth labels are not provided in $\mathcal{T}$, distribution restriction on $\mathcal{T}$ is possible to disturb the procedure of distance optimization on $\mathcal{T}$. While $\mathcal{S}$ has ground truth and the distance optimization on $\mathcal{S}$ is more robust. Thus, $\mathcal{L}_{DIM}$ alleviates the disturbance on distance optimization. 
Huang \textit{et al.}~\cite{huang2019domain} train DNet and ${\Phi}$ with a single objective function to encourange DNet to output a same value with features from both $\mathcal{S}$ and $\mathcal{T}$, \textit{i.e.}, only using Eqn.~\eqref{eqn:dim} to train both DNet and $\Phi$. This leads to a DNet without domain discriminative ability, \textit{i.e.}, outputting a same value for all inputs. Thus, it is not able to supervise $\Phi$ to map samples from $\mathcal{S}$ and $\mathcal{T}$ to a shared feature space. Experiments in Sec.~\ref{sec:model_analysis} shows the advantage of proposed DIM methods over previous methods in~\cite{tzeng2017adversarial} and~\cite{huang2019domain}.

\section{Global-Local Optimization}
\label{sec:glo}

\subsection{Label Prediction}
Unsupervised person ReID is challenging mainly owing to lacking annotations on unlabeled datasets. Before applying distance relationship optimization on $\mathcal{T}$, predicting annotations is an important step that offers guidance for distance optimization. A more precise label prediction leads to a more effective optimization.

In order to predict labels precisely, we first extract features for every image in $\mathcal{T}$ by $\Phi$. Features are then L2-normalized, and the set of features on $\mathcal{T}$ is denoted as $\mathcal{V} = \{v_i| i=1... N_\mathcal{T} \}$. Instead of using Euclidean distance to compute distance matrix, we use the $k$-reciprocal encoding~\cite{zhong2017re} to compute distance among features in $\mathcal{V}$. By additionally considering the distance among neighbors of pairs of features, $k$-reciprocal encoding is able to provide a more precise distance matrix on $\mathcal{V}$.

Based on the distance matrix, we adopt the hierarchical density-based cluster algorithm~\cite{campello2013density} to perform clustering on $\mathcal{V}$. Clustering parameters are set following previous works~\cite{ssg,adaptive-reid}. This separates $\mathcal{V}$ into a set of clusters. In previous methods, all the features in the same cluster are considered to having the same label. However, this involves a number of false positive pairs. To improve the precision of generated positive pairs, we introduce a distance threshold $\alpha$ to discard positive pairs whose distance is larger than $\alpha$. Then, two images in the same cluster and within the distance of $\alpha$ are selected as positive pairs. 

Reducing $\alpha$ improves the precision of generated positive pairs, but also reduces the ratio of recall. We find in practice that the ReID performance is insensitive to $\alpha$ in the range from 0.4 to 0.7. This is because distance distribution is close to a normal distribution with two peaks. Hence, there are few feature pairs whose distances are between 0.4 to 0.7. A threshold large than 0.7 will involve too many false positive pairs, and a threshold smaller than 0.4 will discard too many true positive pairs. We practically set $\alpha$ as 0.5 in experiments. Performance with different values of $\alpha$ is compared in Sec.~\ref{sec:model_analysis}.

Based on generated positive pairs, the annotation matrix ${A}$ can be obtained with ${A}_{i,j} = 1$ denoting the pair of $v_i$ and $v_j$ is positive and vice versa.

\begin{figure}[t]
\begin{center}
\includegraphics[width=0.99\linewidth]{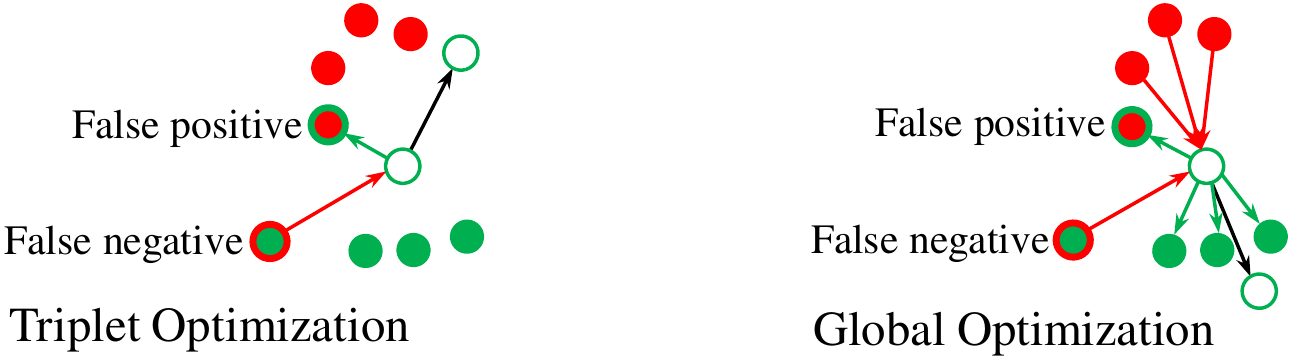}
\end{center}
\vspace{-4mm}
\caption{Comparison of triplet loss and global optimization with noisy labels. Red and greed dots denote negative and positive samples, respectively. Hollow dots denote anchor samples. Red, green, and black arrows denote directions of push, pull and update for anchors, respectively.}
\label{fig:optimization}
\vspace{-4mm}
\end{figure}

\subsection{Global Optimization} 
\label{sec:go}
Based on ${A}$, $\Phi$ can be trained to optimize distance relationships on $\mathcal{V}$. However, as ${A}$ is obtained by unsupervised clustering, it always contains noise. When only several positive/negative pairs are involved in optimization for an anchor in the same time, \textit{e.g.}, triplet loss in previous works, noise will mislead the optimization procedure as illustrated in the left of Fig.~\ref{fig:optimization}. To chase a more precise and effective optimization, we propose the GO method to consider all positive and negative pairs of an anchor in the same time, \textit{i.e.}, globally optimizing distance for the anchor. By involving more samples in the computation of update direction for anchor samples, GO eliminates the effect of noisy label in $A$ and provides precise update direction for anchors as illustrated in the right of Fig.~\ref{fig:optimization}.

For the computation of GO, we propose that GO should satisfy three conditions. 
1): GO should enlarge the similarities of positive pairs and also reduce the similarities of negative pairs for an anchor simultaneously. 
2): GO should be aware of the density of samples around anchors. Some anchors may have dense neighbors, and positive pairs of them should reach rather high similarities. While some other anchors may have sparse neighbors, and restriction on positive pairs of them can be relaxed. Thus, GO should optimize relative similarity of positive and negative samples for each anchor, instead of absolute similarity for each pair. 
3): GO should be aware of the hardness of pairs, \textit{i.e.}, gradients of hard pairs should be larger than gradients of easy pairs. Based on these motivations, we formulate the objective function of GO with respective to the anchor $v_i$ as follows:
\begin{eqnarray}
\label{eqn:go1}
\ell_{GO} ( v_i, \theta ) =  \frac{1}{||A_i||_1} \sum_{j, {A}_{i,j} = 1} \log ( 1 + \frac{S_n}{sim(i,j)}).
\end{eqnarray}
$||A_i||_1$ denotes the number of positive pairs of anchor $v_i$. $sim(i,j)$ denotes the similarity between $v_i$  and $v_j$ computed as:
\begin{eqnarray}
\label{eqn:sim}
sim(i,j) = e^{v_i \cdot v_j / \beta},
\end{eqnarray}
where $\beta$ is a hyper-parameter to adjust the scale of similarity. 
$S_n$ denotes the sum of similarities of negative pairs of $v_i$ as $ S_n = \sum \limits_{k,{A}_{i,k}=0} sim(i,k)$. Note that $sim(i,j)$ denotes the similarity of a positive pair, and $sim(i,k)$ denotes the similarity of a negative pair. 

\textbf{Positive pair weighting.} \ \  
Previous unsupervised optimization methods treat all image pairs from different cameras on the unlabeled dataset with equal weight. However, some camera pairs have little domain gaps, while others have large domain gaps. In person ReID task, matching positive pairs from cameras with large gaps is more challenging and also practically valuable. Thus, model should pay more attention on pulling together positive pairs from cameras with large gaps. To this end, we further propose a positive pair weighting method based on camera gaps to emphasize positive pairs from cameras with large gaps.

We use maximum mean discrepancy to evaluate the gaps among cameras on $\mathcal{T}$. Gap for each camera pair is normalized to the range from 0 to 1 by subtracting the minimum gap and then dividing by the maximum value. We propose to weight positive pairs based on camera gaps linearly, \textit{i.e.}, setting the weight of positive pair $v_i$ and $v_j$ as $g_{i,j}+w$. $g_{i,j}$ denotes the gap between cameras where $v_i$ and $v_j$ are from. And $w$ denotes the base weight for every positive pairs. $w$ is used to enlarge the scale of weights as $g_{i,j}$ is always smaller than 1. Then Eqn.~\eqref{eqn:go1} is re-formulated as:
\begin{eqnarray}
\label{eqn:weight}
\ell_{GO} ( v_i, \theta ) =  \frac{1}{||A_i||_1} \sum_{j,A_{i,j}=1} (g_{i,j}+w) \log ( 1 + \frac{S_n}{sim(i,j)}).
\end{eqnarray}
$w$ is computed based on statistical data. We first compute the mean gap of every positive pairs over $\mathcal{T}$, which is denoted as $g_m$. And then $w$ is set to $1-g_m$ to make the average weight equals to 1.

We select all samples in a training batch as anchors. Then the objective function of GO is formulated as:
\begin{eqnarray}
\mathcal{L}_{GO} ( \theta ) =\! \sum_{i=1}^{K}
\ell_{GO}( v_i, \theta ),
\end{eqnarray}
where $K$ denotes the size of training batch.

\textbf{Discussion.} \ \ 
Proposed objective function in Eqn.~\eqref{eqn:weight} satisfies aforementioned three conditions. The gradient of $\ell_{GO}(v_i,\theta)$ relative to $sim(i,j)$ and $sim(i,k)$ can be respectively computed as:
\begin{eqnarray}
\label{eqn:grad1}
\frac{\partial \ell_{GO}} {\partial sim(i,j)} \cdot \frac{\partial sim(i,j)}{\partial v_i} =  - \frac{(g_{i,j}+w)S_n v_j}{ \beta ||A_i||_1   ( sim(i,j) + S_n )} , 
\end{eqnarray}
\begin{eqnarray}
\label{eqn:grad2}
\frac{\partial \ell_{GO}} {\partial sim(i,k)} \cdot \frac{\partial sim(i,k)}{\partial v_i} = \frac{sim(i,k) v_k }{\beta ||A||_1} \! \!\! \sum_{j,A_{i,j}=1} \!\!\!\! \frac{(g_{i,j}+w)sim(i,j)}{sim(i,j)+S_n}.
\end{eqnarray}
1): As similarities computed by Eqn.~\eqref{eqn:sim} are always larger than 0, gradient in Eqn.~\eqref{eqn:grad1} always pulls $v_i$ close to positive sample $v_j$, and gradient in Eqn.~\eqref{eqn:grad2} always pushes $v_i$ away from negative sample $v_k$. This means minimizing $\ell_{GO}$ leads to enlarging $sim(i,j)$ and reducing $sim(i,k)$ simultaneously. 
2): $\ell_{GO}(v_i,\theta)$ is computed based on the ratio of $S_n$ and $sim(i,j)$, \textit{i.e.}, relative similarity. Thus, $\ell_{GO}(v_i,\theta)$ is aware of the density of neighbors for $v_i$.
3): Moreover, the scale of gradient in Eqn.~\eqref{eqn:grad1} decreases with $sim(i,j)$ increasing, and the scale of gradient in Eqn.~\eqref{eqn:grad2} increases with $sim(i,k)$ increasing. This indicates that $\ell_{GO}(v_i,\theta)$ pays more attention on positive pairs with small similarities, and negative pairs with large similarities. Thus, $\ell_{GO}(v_i,\theta)$ is aware of hardness.

Compared with previous methods, $\mathcal{L}_{GO}$ is more effective at updating model against nosy label as shown in Fig.~\ref{fig:optimization}. Moreover,  $\mathcal{L}_{GO}$ doesn't need any sample mining algorithm and it automatically focuses on hard pairs. Compared with the objective function in~\cite{zhong2019invariance} where enlarging $sim(i,j)$ is conflicting to enlarging similarities of other positive pairs of $v_i$, $\mathcal{L}_{GO}$ avoids the conflict and is more effective at model training.

In model training, extracting all the features on $\mathcal{T}$ to optimize $\mathcal{L}_{GO}$ in each iteration costs a lot of time. Compromising with it, we only extract features of all images in $\mathcal{T}$ at the beginning of training and cache them in a memory bank. In each iteration, features in each training batch are added to corresponding cached features with weight $(100-epoch)/epoch$ to update the memory bank. Updated features in memory bank are L2-normalized. Memory bank costs little memory compared with CNN model, \textit{e.g.}, only around 130 MB on \textit{DukeMTMC-reID}.

Proposed DIM can also be used to bridge gaps between cameras. However, our weighting method shows several advantages: 1) DIM is applied to a pair of domains, thus is expensive to compute for multi-cameras. 2) Weighting method could focus on hard positive samples, thus could better boost feature discriminative power. Experiments have shown the validity of our weighting method.

\subsection{Local Optimization}
\label{sec:lo}
Because of the usage of memory bank, GO uses cached features that are not up-to-date. This will reduce the effectiveness of GO. Thus, we need to optimize distance among those up-to-date features in training batches. On the other hand, in the beginning of the model training, we don't have any prior label information of unlabeled data, and label prediction for GO is not precise due to the weak discriminative ability of the initialized model. Therefore, a label-free training algorithm is needed in the beginning to boost the discriminative ability. Thus, we further propose the LO method performed on up-to-date features in training batches. LO treats all pairs in training batches as negative ones, and the objective function for $v_i$ is formulated as follows:
\begin{eqnarray}
\ell_{LO} (v_i,  \theta ) = \log ( 1 + \sum_{j, j\neq i}^{{K}} sim(i,j)),
\label{eqn:lo_formul}
\end{eqnarray}
Then the objective function of LO in each training batch is:
\begin{eqnarray}
\mathcal{L}_{LO} ( \theta ) =\! \sum_{i=1}^{{K}} \ell_{LO}( v_i, \theta ).
\end{eqnarray}

\textbf{Discussion.} \ \
The gradient of $\ell_{LO}$ relative to $sim(i,k)$ is:
\begin{eqnarray}
\label{eqn:grad_lo}
\frac{\partial \ell_{LO}}{\partial sim(i,k)} \cdot \frac{\partial sim(i,k)}{\partial v_i}= \frac{sim(i,k)v_k}{\beta (1+ \sum_{j, j\neq i}^{{K}} sim(i,j))}.
\end{eqnarray}
Gradient in Eqn.~\eqref{eqn:grad_lo} is always positive and its scale increases with $sim(i,k)$ increasing. This indicates that $\ell_{LO}$ pays more attention on pushing away hard pairs.  It can also be observed that $\mathcal{L}_{LO}$ is more efficient to compute compared with $\mathcal{L}_{GO}$. We also test the method that uses ${A}$ to pull positive pairs in training batches close to each other. However, this method doesn't bring any improvement as shown in Ablation Study in Sec.~\ref{sec:model_analysis}. 
This is because only few positive pairs occur in training batches, and $A$ contains many false positive pairs in the beginning of training.

LO shares certain similarity with Exemplar Invariance (EI) in~\cite{zhong2019invariance}, but it is different in two aspects: 1) Different motivations: LO is computed to enhance our training efficiency. EI is computed with memory bank, thus has similar goal with our GO method. 2) Better efficiency: LO is label-free and is more efficient to compute within a training batch. While, EI relies on memory bank and KNN for positive/negative label prediction. 

\section{Experiment}

\subsection{Dataset}
To evaluate proposed methods, experiments are conducted on three widely used datasets, \textit{i.e.}, \textit{Market1501}~\cite{market}, \textit{DukeMTMC-reID}~\cite{duke}, and \textit{MSMT17}~\cite{msmt}. Details of those datasets are given as follows.

\textit{Market1501} contains 32,668 images of 1,501 identities captured from 6 cameras at Tsinghua University. 12,936 images of 751 identities are selected for training and others are used for testing. In testing set, 3,368 images are selected as query images and remaining 19,732 images are used as gallery images.

\textit{DukeMTMC-reID} contains 36,411 images of 1,812 identities captured from 8 cameras at Duke University. 16,522 images of 702 identities are selected for training and others are used for testing. 3,368 images from testing set are selected as query images, and remaining 19,732 images are used as gallery images.

\textit{MSMT17} contains 126,441 images of 4,101 identities captured from 15 cameras at Peking University. 32,621 images of 1,041 identities are selected for training and others are selected for testing. In testing set, 11,659 images are selected as query images and remaining 82,161 images are used as gallery images.

Following previous works~\cite{market, msmt, duke, zhong2019invariance}, two evaluation metrics, \textit{i.e.}, Cumulative Matching Characteristics (CMC) and mean Average Precision (mAP), are used to evaluate the performance. Rank1, Rank5 and Rank10 accuracies in CMC are reported.

\subsection{Implementation Detail}
Proposed model adopts ResNet50~\cite{resnet} pre-trained on ImageNet~\cite{imagenet} as backbone. The last fully connected layer of ResNet50 is removed and the stride of the last residual block is set to 1. Input images are resized to 256$\times$128. We use random flipping, random cropping, random erasing~\cite{zhong2017random}, and CamStyle~\cite{zhong2018camera} for data augmentation.
In training procedure, each training batch consists of 32 images from source datasets, and 16 images with 3 additional augmented images per image from target datasets. The Adam optimizer is adopted for training. Learning rate is initialized as 0.00035 and decayed by 0.1 every 20 epochs. Model is totally trained for 60 epochs with costing around 5 hours, faster than SSG~\cite{ssg} which costs 2,100 epochs and more than one day. GO is used from the 6-th epoch. Label prediction is performed at the beginning of each epoch on memory bank. Distance threshold $\alpha$ in label prediction is set to 0.5. DNet consists of two fully connected layers with structure of $2048 \times 64$ and $64 \times 1$, and the first fully connected layer is activated by ReLU function. Parameter $\beta$ in Eqn.~\eqref{eqn:sim} is set to 0.05. $\lambda_{LO}$, $\lambda_{GO}$, and $\lambda_{DIM}$ are set as 1, 0.1, and 0.05, respectively. The mean gap of camera pairs on \textit{Market1501}, \textit{DukeMTMC-reID}, and \textit{MSMT17} is 0.60, 0.50, and 0.74, respectively. Then the base weight $w$ for positive pair weighting on corresponding dataset is set as 0.4, 0.5, and 0.26. We use the labeled training set of source domain and the unlabeled training set of target domain to train the model, and use the testing set of target domain to evaluate the model. 

\begin{figure}[t]
\begin{center}
\includegraphics[width=0.99\linewidth]{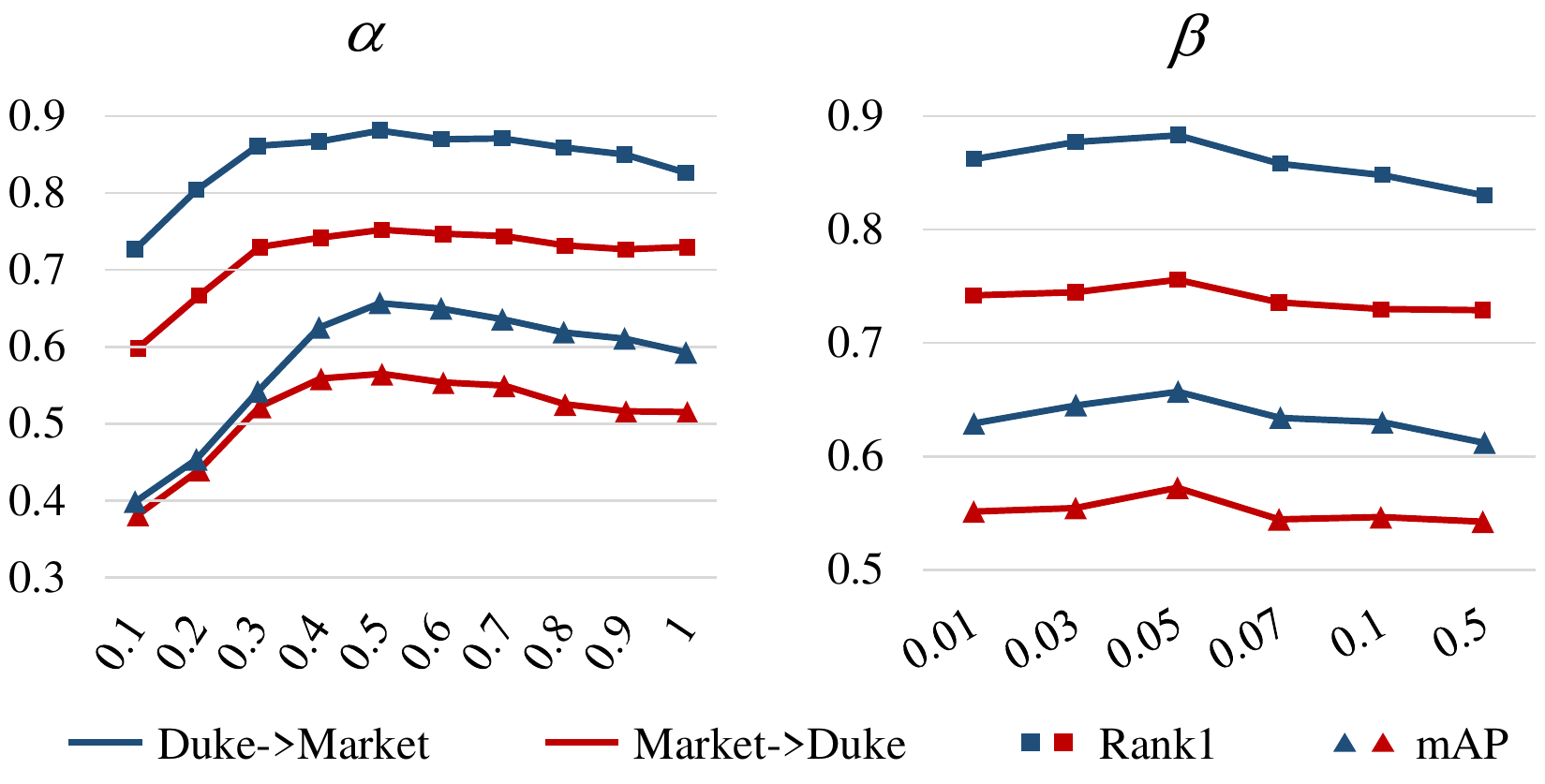}
\end{center}
\vspace{-5mm}
\caption{Evaluation on distance threshold $\alpha$ and temperature parameter $\beta$.}
\label{fig:alpha_t}
\vspace{-4mm}
\end{figure}

\begin{figure}[t]
\begin{center}
\includegraphics[width=0.99\linewidth]{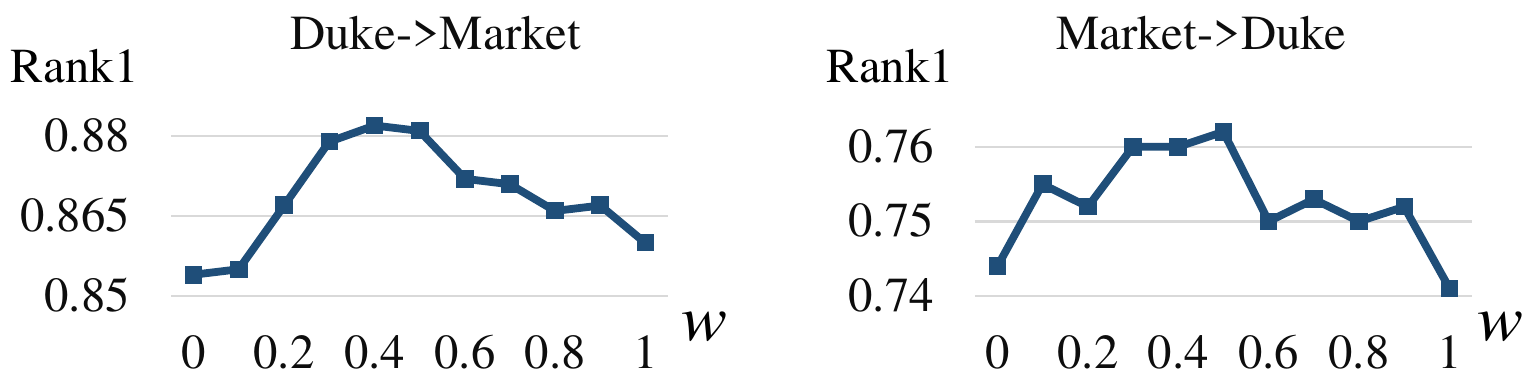}
\end{center}
\vspace{-5mm}
\caption{Evaluation on base weight $w$.
}
\label{fig:w}
\vspace{-4mm}
\end{figure}

\subsection{Model Analysis.}
\label{sec:model_analysis}
In this section, we analysis hyper-parameters on domain adaptive person ReID task. We vary the value of one parameter and keep the others fixed to the optimal values. Experiments show that hyper-parameters selected on one dataset can be applied to other datasets. 

\textbf{Analysis of distance threshold.} \ \
Distance threshold $\alpha$ affects the accuracy and recall rate of selected positive pairs. Evaluation on the effect of $\alpha$ is shown in Fig.~\ref{fig:alpha_t}. It can be observed that performance is insensitive to $\alpha$ in an appropriate range from 0.4 to 0.7. And the best performance is achieved when setting $\alpha$ to 0.5 on both datasets.

\begin{table}[t]
\caption{Comparison on different structures of DNet and training algorithms of DIM. ``D'' and ``M'' denote \textit{DukeMTMC-reID} and \textit{Market-1501}, respectively.}
\vspace{-5mm}
\begin{center}
\setlength{\tabcolsep}{0.43cm}
\resizebox{1.\linewidth}{!}{
\begin{tabular}{l|c|c||c|c}
\hline
\multirow{2}{*}{Structure}  & \multicolumn{2}{c||}{D $\rightarrow$ M} & \multicolumn{2}{c}{M $\rightarrow$ D} \\
\cline{2-5}
 &  Rank1 & mAP & Rank1 & mAP  \\
\hline
Without DIM & 0.865 & 0.610 & 0.727 & 0.541\\
\hline
$2048 \times 1$ & 0.876 & 0.619 & 0.73 & 0.534 \\
$2048 \times 64 \times1$ & \textbf{0.883} & \textbf{0.651} &\textbf{0.762} &\textbf{0.583}\\
$2048 \times 128 \times64 \times 1$ & 0.874 & 0.611& {0.729} & {0.543} \\
\hline
\hline
ADDA~\cite{tzeng2017adversarial} &0.872 & 0.632& {0.733} & {0.551} \\
OCC~\cite{huang2019domain} &0.866 & 0.623& {0.718} & {0.530} \\
DIM & \textbf{0.883} & \textbf{0.651} &\textbf{0.762} &\textbf{0.583}\\
\hline
\end{tabular}}
\end{center}
\label{tab:abalation-adversarial}
\vspace{-3mm}
\end{table}

\begin{table}[t]
\caption{Evaluation of loss weights. ``D'' and ``M'' denote \textit{DukeMTMC-reID} and \textit{Market-1501}, respectively.}
\vspace{-5mm}
\begin{center}
\resizebox{1.\linewidth}{!}{
\begin{tabular}{p{7mm}|p{7mm}|p{15mm}<{\centering}|p{15mm}<{\centering}||p{15mm}<{\centering}|p{15mm}<{\centering}}
\hline
\multicolumn{2}{c|}{\multirow{2}{*}{Weight}}  & \multicolumn{2}{c||}{D $\rightarrow$ M}  & \multicolumn{2}{c}{M $\rightarrow$ D} \\
\cline{3-6}
 \multicolumn{2}{c|}{} & Rank1 & mAP & Rank1 & mAP  \\

\hline
\multicolumn{2}{c|}{Directly transfer} &0.530 & 0.224 &0.349&0.171\\
\hline
\multirow{5}{*}{$\lambda_{GO}$}
& 0& 0.693 & 0.313 & 0.571 & 0.315 \\
&0.01& 0.801 & 0.617 & 0.710 & 0.516 \\
&0.1&\textbf{0.883} &\textbf{ 0.651} & \textbf{0.762} & \textbf{0.583} \\
&1& 0.857 & 0.624 & 0.727 & 0.544 \\
&5 &0.810 & 0.601 & 0.712 & 0.539 \\
\hline

\hline
\multirow{5}{1mm}{\raggedleft $\lambda_{LO}$}
& 0& 0.840 & 0.612 & 0.721 & 0.538 \\
&0.01& 0.866 & 0.632 & 0.744 & 0.557 \\
&0.1& 0.871 & 0.649 & 0.745 & 0.561 \\
&1& \textbf{0.883} &\textbf{ 0.651} & \textbf{0.762} & \textbf{0.583} \\
&5 &0.811 & 0.601 & 0.718 & 0.529 \\
\hline
\hline
\multirow{5}{*}{$\lambda_{DIM}$}
& 0& 0.865 & 0.610 & 0.727 & 0.541 \\
&0.01& 0.871 & 0.621 & 0.733 & 0.540 \\
&0.05&\textbf{0.883} &\textbf{ 0.651} & \textbf{0.762} & \textbf{0.583} \\
&0.1& 0.865 & 0.581 & 0.733 & 0.540 \\
&0.5& 0.817 & 0.479 & 0.697 & 0.485 \\
\hline
\end{tabular}}
\end{center}
\label{tab:abalation-weights}
\vspace{-6mm}
\end{table}

\textbf{Analysis of temperature parameter.} \ \
Temperature parameter $\beta$ affects the computation of $\mathcal{L}_{GLO}$.
Evaluation on the effect of $\beta$ is also shown in Fig.~\ref{fig:alpha_t}. It can be observed that the best performance is achieved when $\beta$ is set to 0.05 on both datasets.

\textbf{Analysis of base weight.} \ \
In positive pair weighting, base weight $w$ is computed as $1-g_m$.
Different value of $w$ is evaluated as shown in Fig.~\ref{fig:w}. It can be observed that setting $w$ to 0.4 and 0.5 on \textit{Market1501} and \textit{DukeMTMC-reID} based on $1-g_m$ achieves the best performance. This indicates the validity of the proposed method to compute base weight $w$ for positive pair weighting.

\textbf{Analysis of DIM.} \ \
Different structures of discriminator and different learning algorithms are compared with proposed DIM method in this section. Experimental results are summarized in Table~\ref{tab:abalation-adversarial}. It can be observed that, DIM with different structures can always improve the performance on both datasets.
This indicates the validity of DIM to narrow down domain gap in feature space and improve the discriminative ability on the target dataset.
DNet with shallow structure is not effective at eliminating domain gap because of lacking domain discriminative ability. And DNet with deep structure provides an unnecessary strong restriction on feature extractor and thus disturbs the distance optimization. Comparison among three structures shows that using a two-layer network achieves the best performance. Different learning algorithms, \textit{i.e.}, ADDA~\cite{tzeng2017adversarial} and OCC~\cite{huang2019domain}, are also compared in Table~\ref{tab:abalation-adversarial}.
It can be observed that DIM outperforms ADDA and OCC on both datasets, \textit{e.g.}, by 1.9\% and 2.8\% in mAP on \textit{Market-1501}.

\textbf{Analysis of loss weights.} \ \
We compare different values of loss weights for $\lambda_{GO}$, $\lambda_{LO}$, and $\lambda_{DIM}$. The results are summarized in Table~\ref{tab:abalation-weights}. Compared with setting $\lambda_{GO}$ to 0, setting it larger than 0 largely improves the performance, and the best performance is achieved with $\lambda_{GO}$ as 0.1. This demonstrates that GO is effective at enhancing the discriminative ability on target domain. Compared with setting $\lambda_{LO}$ to 0, setting it to the range from 0.01 to 1 achieves clear improvement and setting it to 1 achieves the best performance. Setting $\lambda_{DIM}$ larger than 0 shows improvements against setting it to 0, and setting it to 0.05 achieves the best performance.

\begin{table*}[t]
\caption{Performance comparison with state-of-the-art methods on \textit{Market1501} and \textit{DukeMTMC-reID}. ``\textit{Market}'' and ``\textit{Duke}'' denote \textit{Market1501} and \textit{DukeMTMC-reID}, respectively. ``$^{\dagger}$'' denotes that methods are semi-supervised. ``$^{*}$'' denotes temporal information is used. And ``$^{\ddagger}$'' denotes multi-scale features are used.}
\vspace{-5mm}
\begin{center}
\setlength{\tabcolsep}{0.35cm}
\resizebox{1.\linewidth}{!}{
\begin{tabular}{l|c|c|c|c|c|c||c|c|c|c|c}
\hline
\multirow{2}{*}{Method} & \multirow{2}{*}{Reference} & \multicolumn{5}{c||}{Market-1501} & \multicolumn{5}{c}{DukeMTMC-reID}\\
\cline{3-12}
 && Source & Rank1 & Rank5 & Rank10 & mAP  & Source & Rank1 & Rank5 & Rank10 & mAP\\
\hline
Supervised & Baseline
&\textit{Market} & 0.899& 0.969 & 0.981 & 0.728
&\textit{Duke} & 0.770 & 0.895 & 0.921 &0.616\\
\hline
LOMO~\cite{lomo} & CVPR'15
&None&0.272&0.416&0.491&0.080
&None&0.123&0.213&0.266&0.048\\
BOW~\cite{market}& ICCV'15
&None&0.358&0.524&0.603&0.148
&None&0.171&0.288&0.349&0.083\\
DBC~\cite{ding2019towards} & BMVC'19
&None&0.692&0.830&0.878&0.413
&None&0.515&0.646&0.701&0.300\\
BUC~\cite{lin2019bottom} & AAAI'19
&None&0.662&0.796&0.845&0.383
&None&0.474&0.626&0.684&0.275\\
GLO & This paper
&None&\textbf{0.774}&\textbf{0.880}&\textbf{0.901}&\textbf{0.457}
&None&\textbf{0.605}&\textbf{0.722}&\textbf{0.757}&\textbf{0.364}\\
\hline
\hline
TAUDL$^{\dagger}$~\cite{li2018unsupervised} & ECCV'18
&Tracklet&0.637&-&-&0.412
&Tracklet&0.617&-&-&0.435\\
UTAL$^{\dagger}$~\cite{li2019unsupervised-pami} & PAMI'19
&Tracklet&0.692&-&-&0.462
&Tracklet&0.623&-&-&0.446\\
MAR~\cite{yu2019unsupervised}&CVPR'19
&\textit{MSMT17}&0.677&0.819&-&0.400
&\textit{MSMT17}&0.671&0.798&-&0.480\\
PAUL~\cite{yang2019patch}& CVPR'19
&\textit{MSMT17}&0.685&0.824&0.874&0.401
&\textit{MSMT17}&0.720&0.827&0.860&0.532\\
CASCL~\cite{wu2019unsupervised}&ICCV'19
&\textit{MSMT17}&0.654&0.806&0.862&0.355
&\textit{MSMT17}&0.593&0.732&0.778&0.378\\
UDA~\cite{adaptive-reid} & arXiv
&\textit{Duke}&0.758&0.895&0.932&0.537
&\textit{Market}&0.684&0.801&0.835&0.490\\
GPP~\cite{zhong2019learning} & arXiv
&\textit{Duke}& 0.841&0.928&0.954&0.638
&\textit{Market}&0.740&0.837&0.874&0.544\\
HHL~\cite{zhong2018generalizing} & ECCV'18
&\textit{Duke}& 0.622&0.788&0.840&0.314
&\textit{Market}&0.469&0.610&0.667&0.272\\
ECN~\cite{zhong2019invariance} & CVPR'19
&\textit{Duke}& 0.751&0.876&0.916&0.430
&\textit{Market}&0.633&0.758&0.804&0.404\\
PAUL~\cite{yang2019patch}& CVPR'19
&\textit{Duke}& 0.667&-&-&0.368
&\textit{Market}&0.561&-&-&0.357\\
ATNet~\cite{Liu_2019_CVPR}&CVPR'19
&\textit{Duke}&0.557&0.732&0.794&0.256
&\textit{Market}&0.451&0.595&0.642&0.249\\
CR\_GAN\cite{Chen_2019_ICCV} & ICCV'19
&\textit{Duke}& 0.777&0.897&0.927&0.540
&\textit{Market}&0.689&0.802&0.847&0.486\\
SSG$^{\ddagger}$~\cite{ssg} & ICCV'19
&\textit{Duke}&0.800&0.900&0.924&0.583
&\textit{Market}&0.730&0.806&0.832&0.534\\
DA\_2S~\cite{huang2019sbsgan}&ICCV'19
&\textit{Duke}&0.585&-&-&0.273
&\textit{Market}&0.535&-&-&0.308\\
CAL$^{*\dagger}$~\cite{Qi_2019_ICCV}&ICCV'19
&\textit{Duke}&0.737&-&-&0.496
&\textit{Market}&0.640&-&-&0.456\\
PDA-Net~\cite{Li_2019_ICCV}&ICCV'19
&\textit{Duke}&0.752&0.863&0.902&0.476
&\textit{Market}&0.632&0.770&0.825&0.451\\
PAST$^{\ddagger}$~\cite{Zhang_2019_ICCV}&ICCV'19
&\textit{Duke}&0.784&-&-&0.546
&\textit{Market}&0.724&-&-&0.543\\
\hline
DIM+GLO &This paper
&\textit{Duke}&\textbf{0.883}&\textbf{0.947}&\textbf{0.963}&\textbf{0.651}
&\textit{Market}&\textbf{0.762}&\textbf{0.857}&\textbf{0.885}&\textbf{0.583}\\
\hline
\end{tabular}}
\end{center}
\label{tab:compared-others}
\vspace{-2mm}
\end{table*}

\begin{table}[t]
\newcommand{\tabincell}[2]{\begin{tabular}{@{}#1@{}}#2\end{tabular}}
\centering
\caption{Performance comparison with state-of-the-art methods on \textit{MSMT17}.}
\vspace{-5mm}
\label{tab:compared-msmt}
\setlength{\tabcolsep}{0.3cm}
\resizebox{1.\linewidth}{!}{
\begin{tabular}{l|c|c|c|c|c}
\hline
\multirow{2}{*}{Method} & \multirow{2}{*}{Reference} & \multicolumn{4}{c}{MSMT17} \\
\cline{3-6}
 && Source & Rank1
 & Rank10 & mAP  \\
\hline
PTGAN~\cite{msmt} & CVPR'18 & \multirow{6}{*}{\tabincell{c}{\textit{Market-}\\ \textit{1501}}} & 0.102
&0.244&0.029\\
ECN~\cite{zhong2019invariance}& CVPR'19 & &0.253
&0.421&0.085\\
GPP~\cite{zhong2019learning} & arXiv & &0.404
&0.587&0.152\\
SSG~\cite{ssg} & ICCV'19 & &0.316 & 0.496 & 0.132\\
SSG++~\cite{ssg} & ICCV'19& &0.376 & 0.572 & 0.166\\
DIM+GLO &This paper & &\textbf{0.497}
&\textbf{0.661}&\textbf{0.207}\\
\hline
\hline
PTGAN~\cite{msmt} & CVPR'18 & \multirow{6}{*}{\tabincell{c}{\textit{DukeMT}\\ \textit{MC-reID}}} &0.118
&0.274&0.033\\
ECN~\cite{zhong2019invariance}& CVPR'19 & &0.302
&0.468&0.102\\
GPP~\cite{zhong2019learning} & arXiv & &0.425
&0.615&0.160\\
SSG~\cite{ssg} & ICCV'19 & &0.322 & 0.512 & 0.133\\
SSG++~\cite{ssg} & ICCV'19& &0.416 & 0.622 & 0.183\\
DIM+GLO &This paper & &\textbf{0.565}
&\textbf{0.700}&\textbf{0.244}\\
\hline
\end{tabular}}
\vspace{-7mm}
\end{table}

\begin{figure}[t]
\begin{center}
\includegraphics[width=0.99\linewidth]{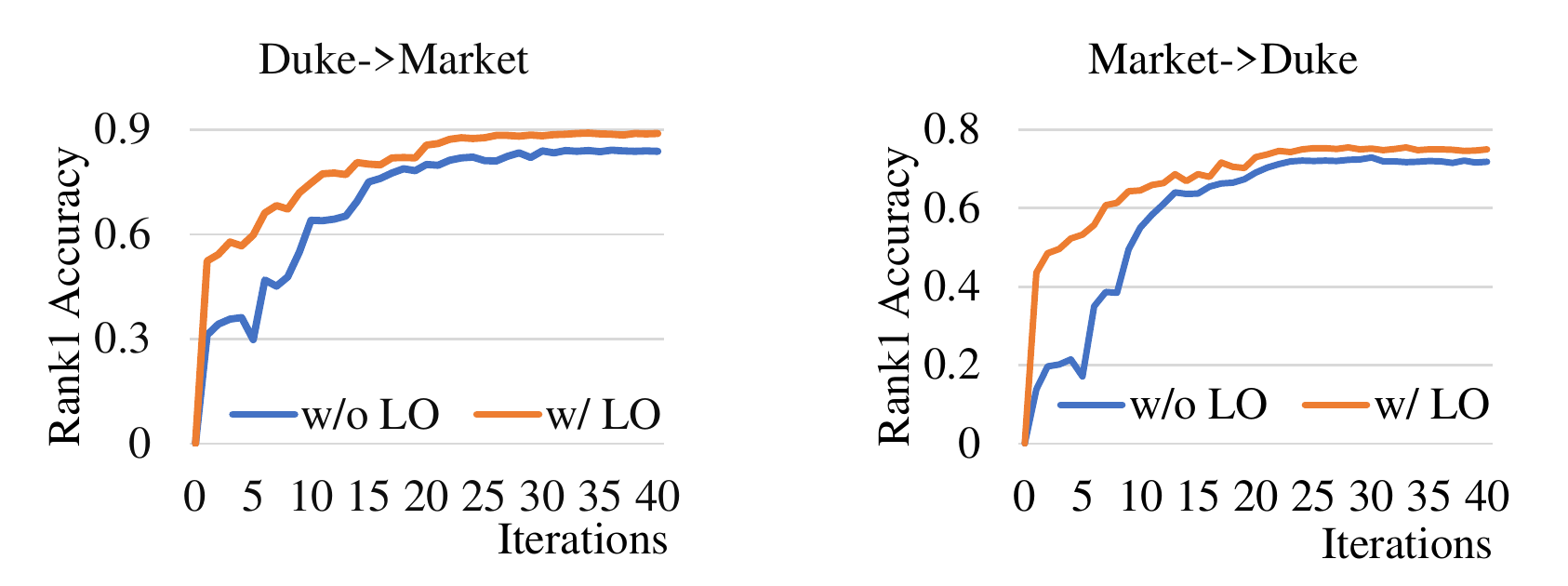}
\end{center}
\vspace{-6mm}
\caption{Comparison of training procedures with and without LO.}
\label{fig:curve_lo}
\vspace{-4mm}
\end{figure}

\textbf{Ablation Study.} \ \
Each component of our methods is evaluated in this section and experimental results can be observed in Table~\ref{tab:abalation-weights} with setting $\mathcal{L}_{GO}$, $\mathcal{L}_{LO}$, and $\mathcal{L}_{DIM}$ to 0, respectively. It can be observed that, compared with directly transferring model trained on source datasets to target datasets, DIM+GLO boosts the performance by a large margin, \textit{e.g.}, improving mAP by 42.7\% on \textit{Market-1501}. Compared with the model without $\mathcal{L}_{GO}$, DIM+GLO largely improves mAP by 33.8\% and 26.8\% on \textit{Market-1501} and \textit{DukeMTMC-reID}, respectively. This demonstrates that proposed GO can effectively improve the discriminative ability of learned features. It can also be observed that $\mathcal{L}_{LO}$ and $\mathcal{L}_{DIM}$ improves the performance clearly, \textit{e.g.}, improves Rank1 accuracy on \textit{DukeMTMC-reID} by 4.1\% and 3.5\%, respectively. This indicates the validity of LO and DIM methods. When we use annotation to pull positive pairs close to each other in LO as discussed in Sec.~\ref{sec:lo}, the Rank1 accuracy on \textit{Market-1501} and \textit{DukeMTMC-reID} is 0.873 and 0.749, showing no improvement. This indicates LO doesn't need annotation. We compare the training procedures with and without LO in Fig.~\ref{fig:curve_lo}. It can be observed that LO largely boosts the performance in the beginning of training, and also improves the final performance.

\subsection{Comparison with State-of-the-art Methods}
\label{sec:experiment_comparison}
In this section, proposed methods are compared with many state-of-the-art methods. The comparison results on \textit{Market-1501} and \textit{DukeMTMC-reID} are summarized in Table~\ref{tab:compared-others}.
And the comparison results on \textit{MSMT17} are shown in Table~\ref{tab:compared-msmt}.

\textbf{Comparison on the unsupervised person ReID.} \ \
For unsupervised person ReID task, source dataset is not available. Therefore, we only use images from target datasets and $\mathcal{L}_{GLO}$ in model training. Comparisons are performed on \textit{Market-1501} and \textit{DukeMTMC-reID}. Compared methods include LOMO~\cite{lomo}, BOW~\cite{market}, DBC~\cite{ding2019towards}, and BUC~\cite{lin2019bottom}. It can be observed from Table~\ref{tab:compared-others} that proposed GLO achieves the best performance on both datasets and outperforms state-of-the-art methods by a clear margin. For example, GLO outperforms DBC, currently the best unsupervised person ReID method, by 8.2\% and 9\% in Rank1 accuracy on \textit{Market-1501} and \textit{DukeMTMC-reID}, respectively. As DBC also predicts labels via unsupervised clustering, these results indicate that proposed GLO method can effectively improve the discriminative ability of learned features. It can also be observed that GLO also outperforms several domain adaptive methods such as  PAUL~\cite{yang2019patch} and DA\_2S~\cite{huang2019sbsgan}. Specially, compared with ECN~\cite{zhong2019invariance} that uses a source dataset, GLO still outperforms it by 2.7\% in mAP on \textit{Market-1501}. As ECN uses the same data augmentation method and a similar optimization method, the compared results further demonstrate the advantage of proposed GLO method. 

\textbf{Comparison on domain adaptive person ReID on \textit{Market-1501} and \textit{DukeMTMC-reID}.} \ \ 
Proposed DIM+GLO model is compared with recent state-of-the-art methods including TAUDL$^{\dagger}$~\cite{li2018unsupervised}, UTAL$^{\dagger}$~\cite{li2019unsupervised-pami}, MAR~\cite{yu2019unsupervised}, PAUL~\cite{yang2019patch}, CASCL~\cite{wu2019unsupervised}, UDA~\cite{adaptive-reid}, GPP~\cite{zhong2019learning}, HHL~\cite{zhong2018generalizing}, ECN~\cite{zhong2019invariance}, ATNet~\cite{Liu_2019_CVPR}, CR\_GAN\cite{Chen_2019_ICCV}, SSG$^\ddagger$~\cite{ssg}, DA\_2S~\cite{huang2019sbsgan}, CAL$^{*\dagger}$~\cite{Qi_2019_ICCV}, PDA-Net~\cite{Li_2019_ICCV}, and PAST$^{\ddagger}$~\cite{Zhang_2019_ICCV}. The comparison results are summarized in Table~\ref{tab:compared-others}. It can be observed that proposed DIM+GLO model outperforms state-of-the-art methods on both datasets. Specially, DIM+GLO outperforms SSG$^\ddagger$ and PAST$^\ddagger$ that use multi-scale features as enhancement and similar label prediction methods by a clear margin, \textit{e.g.,} 8.3\% and 9.9\% in Rank1 accuracy on \textit{Market-1501}. DIM+GLO also outperforms three semi-supervised methods, \textit{i.e.,} TAUDL$^\dagger$, UTAL$^\dagger$, and CAL$^{*\dagger}$. For instance, CAL$^{*\dagger}$ additionally uses temporal information and DIM+GLO still outperforms it by 14.6\% and 12.2\% in Rank1 accuracy on \textit{Market-1501} and \textit{DukeMTMC-reID}, respectively. It can also be observed that DIM+GLO achieves comparable performance with supervised baseline models, \textit{e.g.}, only 1.6\% and 0.8\% lower in Rank1 accuracy on \textit{Marekt-1501} and \textit{DukeMTMC-reID}, respectively. This indicates that proposed DIM+GLO is powerful to learn discriminative person features without annotation.

\textbf{Comparison on domain adaptive person ReID on \textit{MSMT17}.}
\textit{MSMT17} is currently the largest person ReID dataset, which is more challenging than \textit{DukeMTMC-reID} and \textit{Market-1501}. We compare proposed domain adaptive model DIM+GLO with state-of-the-art methods including ECN~\cite{zhong2019invariance}, SSG~\cite{ssg}, SSG++~\cite{ssg}, and GPP~\cite{zhong2019learning}. Performance comparison is shown in Table~\ref{tab:compared-msmt}. It can be observed that proposed DIM+GLO outperforms previous methods by a large margin. For example, DIM+GLO achieves \textbf{0.497} and \textbf{0.565} in Rank1 accuracy when using \textit{Market-1501} and \textit{DukeMTMC-reID} as the  source dataset, respectively. And both of the accuracies outperform the best previous accuracy 0.425 achieve by GPP~\cite{zhong2019learning} with \textit{DukeMTMC-reID} as the labeled dataset by \textbf{7.2\%} and \textbf{14\%}, respectively. SSG++ is a semi-supervised method and proposed DIM+GLO model also outperforms it by a large margin. This further demonstrates the effectiveness of proposed method on large-scale person ReID task.

\section{Conclusion}
This paper proposes a coupling optimization method for domain adaptive person ReID consisting of the Domain-Invariant Mapping (DIM) method for knowledge transfer and the Global-Local distance Optimization (GLO) method for unsupervised model training. DIM encourages the model to map images from both labeled and unlabeled domains into a shared domain in feature space, enhancing the efficiency of knowledge transfer. GLO involves more samples in distance optimization, enhancing the robustness to noisy label prediction on unlabeled datasets. Experiments on three large-scale datasets show that our methods outperform state-of-the-art methods by a clear margin. Specially, proposed unsupervised training method even outperforms several recent domain adaptive methods.

\begin{acks}
This work is supported in part by Peng Cheng Laboratory, The National Key Research and Development Program of China under Grant No. 2018YFE0118400, in part by Beijing Natural Science Foundation under Grant No. JQ18012, in part by Natural Science Foundation of China under Grant No. 61936011, 61620106009, 61425025, 61572050, 91538111.
\end{acks}

\balance
\bibliographystyle{ACM-Reference-Format}
\bibliography{sample-sigconf}

\end{document}